# Composition of Probability Measures on Finite Spaces

Radim Jiroušek*
Laboratory of Intelligent Systems
University of Economics, Prague
and
Institute of Information Theory and Automation
Academy of Sciences of the Czech Republic

**Abstract**

Decomposable models and Bayesian networks can be defined as sequences of oligo-dimensional probability measures connected with *operators of composition*. The preliminary results suggest that the probabilistic models allowing for effective computational procedures are represented by sequences possessing a special property; we shall call them *perfect* sequences.

The present paper lays down the elementary foundation necessary for further study of iterative application of operators of composition. We believe to develop a technique describing several graph models in a unifying way. We are convinced that practically all theoretical results and procedures connected with decomposable models and Bayesian networks can be translated into the terminology introduced in this paper. For example, complexity of computational procedures in these models is closely dependent on possibility to change the ordering of oligo-dimensional measures defining the model. Therefore, in this paper, lot of attention is paid to possibility to change ordering of the operators of composition.

## 1 NOTATION

Let $\{X_i\}_{i \in V}$ be a finite system of finite sets. We shall deal with probability measures on the Cartesian product space

$$\times_{i \in V} X_i = X_V$$

and its subspaces

$$\times_{i \in L} X_i = X_L$$

for $L \subset V$.

Consider $K \subseteq L \subseteq V$ and a probability measure $P$ defined on $X_K$. By $\Pi^{(L)}$ we shall denote the set of all probability measures defined on $X_L$. Similarly, $\Pi^{(L)}(P)$ will denote the system of all extensions of the measure $P$ to measures on $X_L$:

$$\Pi^{(L)}(P) = \left\{ Q \in \Pi^{(L)} : Q^{(K)} = P \right\},$$

where $Q^{(K)}$ denote the marginal measure of the measure $Q$ on $X_K$.

Now, let us introduce an operator $\triangleright$ of *composition*. To make it clear from the very beginning, let us stress that it is just a generalization of the idea of computing the three-dimensional distribution from two two-dimensional ones introducing the conditional independence:

$$\begin{aligned} P(X_1, X_2) \triangleright Q(X_2, X_3) &= \frac{P(X_1, X_2) Q(X_2, X_3)}{Q(X_2)} \\ &= P(X_1, X_2) Q(X_3 | X_2). \end{aligned}$$

Consider two probability measures $P \in \Pi^{(J)}$ and $Q \in \Pi^{(K)}$, such that $Q^{(J \cap K)}$ dominates[1] $P^{(J \cap K)}$; in symbol:

$$P^{(J \cap K)} \ll Q^{(J \cap K)}.$$

$P^{(J \cap K)}$, $Q^{(J \cap K)}$ denote again the marginal measures of $P, Q$ respectively, on $X_{J \cap K}$. The (right) composition of these two measures is defined by the formula

$$P \triangleright Q = \frac{PQ}{Q^{(J \cap K)}}.$$

---

[1]The concept of dominance (or absolute continuity) $P \ll Q$ on finite space $X$ simplifies to

$$\forall x \in X \quad (Q(x) = 0 \Longrightarrow P(x) = 0).$$

*E-mail: radim@vse.cz

Since we assume $P^{(J \cap K)} \ll Q^{(J \cap K)}$, if for any $x \in \mathbf{X}_{(J \cap K)}$ $Q^{(J \cap K)}(x) = 0$ then there is a product of two zeros in the nominator and we take, quite naturally,

$$\frac{0.0}{0} = 0.$$

If $J \cap K = \emptyset$ then $Q^{(J \cap K)} = 1$ and the formula degenerates to a simple product of $P$ and $Q$.

It is obvious from this definition that $P \triangleright Q$ is a probability measure from $\Pi^{(J \cup K)}$ and that its marginal measure $(P \triangleright Q)^{(J)}$ equals $P$. In case $P^{(J \cap K)} \not\ll Q^{(J \cap K)}$ the expression $P \triangleright Q$ remains undefined.

**Example**

Let us illustrate by a simple example difficulties which can occur when $P^{(J \cap K)} \not\ll Q^{(K \cap J)}$.

Consider the measures $P \in \Pi^{(\{1,2\})}$ and $Q \in \Pi^{(\{2,3\})}$ given in the following tables.

Table 1: Prob. Measure $P$

| $P$ | $x_1 = 0$ | $x_1 = 1$ |
|---|---|---|
| $x_2 = 0$ | $\frac{1}{2}$ | $\frac{1}{2}$ |
| $x_2 = 1$ | 0 | 0 |

Table 2: Prob. Measure $Q$

| $Q$ | $x_3 = 0$ | $x_3 = 1$ |
|---|---|---|
| $x_2 = 0$ | 0 | 0 |
| $x_2 = 1$ | $\frac{1}{2}$ | $\frac{1}{2}$ |

The reader can easily see that for any $x \in \mathbf{X}_{\{1,2,3\}}$, $x = (x_1, x_2, x_3)$

$$P(x_1, x_2) Q(x_2, x_3) = 0$$

since for $x_2 = 1$ $P(x_2, x_3) = 0$ and for $x_2 = 0$ $Q(x_1, x_2) = 0$. $\diamond$

To avoid necessity of writing brackets distinguishing expression $((P \triangleright Q) \triangleright R)$ from $(P \triangleright (Q \triangleright R))$ we use also an analogous operator of left composition $\triangleleft$

$$P \triangleleft Q = \frac{PQ}{P^{(J \cap K)}}$$

which is properly defined whenever $Q^{(J \cap K)} \ll P^{(J \cap K)}$, otherwise it is again undefined. Hereafter we shall always assume that the operators are applied from left to right.

## 2   BASIC PROPERTIES

**Lemma 1.** Let $K \subseteq L \subseteq V$. For any probability measures $P \in \Pi^{(K)}$ and $Q \in \Pi^{(L)}$ such that $P \ll Q^{(K)}$,

$$P \triangleright Q \ll Q$$

and for any $R \in \Pi^{(L)}(P)$

$$R \ll Q \Longrightarrow R \ll P \triangleright Q.$$

*Proof.*
The assertion directly follows from the definition of the operator $\triangleright$ which can be for the current situation written

$$P \triangleright Q = \frac{PQ}{Q^{(K)}}.$$

From this formula it follows evidently that for any $x \in \mathbf{X}_L$

$$Q(x) = 0 \Longrightarrow (P \triangleright Q)(x) = 0,$$

which proves that $P \triangleright Q \ll Q$.

Analogously, let $R \in \Pi^{(L)}(P)$ be dominated by $Q$. Consider an $x \in \mathbf{X}_L$ for which

$$(P \triangleright Q)(x) = \frac{P(x^{(K)}) Q(x)}{Q^{(K)}(x^{(K)})} = 0,$$

where $x^{(K)}$ for $x \in \mathbf{X}_L$ denote the projection of $x$ into the subspace $\mathbf{X}_K$. That means that either $P(x^{(K)}) = 0$ or $Q(x) = 0$ (or both). If $P(x^{(K)}) = 0$ then $R^{(K)}(x^{(K)}) = 0$ as $R^{(K)}$ equals $P$ since $R \in \Pi^L(P)$. Therefore also $R(x) = 0$. On the other hand, if $Q(x) = 0$ then $R(x) = 0$ because $R$ is dominated by $Q$. □

In the proofs, we shall often compute a marginal measure from a measure defined as a composition of two (or several) oligo-dimensional measures. Therefore, it is important to realize that generally

$$(P \triangleright Q)^{(M)} \neq P^{(K \cap M)} \triangleright Q^{(L \cap M)}.$$

As a simple example can serve

$$(P(X_1, X_2) \triangleright Q(X_2, X_3))^{(\{1,3\})}$$
$$\neq (P(X_1, X_2))^{(\{1\})} \triangleright (P(X_2, X_3))^{(\{3\})}$$
$$= P(X_1) Q(X_3).$$

Nevertheless, the following simple assertion presents a sufficient condition under which the equality in the above expression holds.

**Lemma 2.** Let $K, L, M \subseteq V$. If $M \supseteq K \cap L$ than for any probability measures $P \in \Pi^{(K)}$ and $Q \in \Pi^{(L)}$

$$(P \triangleright Q)^{(M)} = P^{(K \cap M)} \triangleright Q^{(L \cap M)}.$$

*Proof.*
To prove this, it is enough to marginalize $(P \triangleright Q)$ as, for example, in the following way.

$$\begin{aligned}(P \triangleright Q)^{(M)} &= ((P \triangleright Q)^{(K \cup M)})^{(M)} \\ &= (P \triangleright Q^{(L \cap M)})^{(M)} \\ &= P^{(K \cap M)} \triangleright Q^{(L \cap M)}.\end{aligned}$$

□

## 3  ITERATIONS OF THE OPERATORS OF COMPOSITION

We shall say that a pair of measures $P_1 \in \Pi^{(K_1)}, P_2 \in \Pi^{(K_2)}$ is *consistent* if

$$P_1^{(K_1 \cap K_2)} = P_2^{(K_1 \cap K_2)}.$$

Directly from the definition of the operators $\triangleleft$ and $\triangleright$ we get the following trivial assertion.

**Lemma 3.** Let for $P_1 \in \Pi^{(K_1)}$ and $P_2 \in \Pi^{(K_2)}$ either $P_1^{(K_1 \cap K_2)} \ll P_2^{(K_1 \cap K_2)}$ or $P_2^{(K_1 \cap K_2)} \ll P_1^{(K_1 \cap K_2)}$. $P_1$ and $P_2$ are consistent if and only if

$$P_1 \triangleright P_2 = P_1 \triangleleft P_2.$$

□

The situation becomes somewhat more complicated when the operators are applied (at least) twice. It is clear that generally[2]

$$\begin{aligned}P_1 \triangleright P_2 \triangleright P_3 &\neq P_1 \triangleright (P_2 \triangleright P_3) = P_2 \triangleright P_3 \triangleleft P_1, \\ P_1 \triangleright P_2 \triangleright P_3 &\neq P_1 \triangleright P_3 \triangleright P_2, \\ P_1 \triangleright P_2 \triangleright P_3 &\neq P_1 \triangleleft P_2 \triangleleft P_3, \\ P_1 \triangleright P_2 \triangleright P_3 &\neq P_3 \triangleleft P_2 \triangleleft P_1, \\ P_1 \triangleleft P_2 \triangleleft P_3 &\neq P_1 \triangleleft P_3 \triangleleft P_2.\end{aligned}$$

Now, let us present a few lemmata saying under which conditions some of these equalities hold. Whenever in this section we shall use probability measures $P_1, P_2, P_3$, we shall assume that $P_i \in \Pi^{(K_i)}$ for $i = 1, 2, 3$.

**Lemma 4.** If $K_1 \supseteq (K_2 \cap K_3)$ then

$$P_1 \triangleright P_2 \triangleright P_3 = P_1 \triangleright P_3 \triangleright P_2$$

---
[2] As said above, operators $\triangleright$ and $\triangleleft$ are always applied from left to right.

*Proof.*
Under the assumption $K_1 \supseteq (K_2 \cap K_3)$

$$K_3 \cap (K_1 \cup K_2) = K_3 \cap K_1,$$

$$K_2 \cap (K_1 \cup K_3) = K_2 \cap K_1,$$

and therefore the expressions

$$P_1 \triangleright P_2 \triangleright P_3 = \frac{P_1 P_2 P_3}{P_2^{(K_1 \cap K_2)} P_3^{(K_3 \cap (K_1 \cup K_2))}},$$

$$P_1 \triangleright P_3 \triangleright P_2 = \frac{P_1 P_2 P_3}{P_3^{(K_1 \cap K_3)} P_2^{(K_2 \cap (K_1 \cup K_3))}}$$

are equivalent each to other.

□

The next assertion expresses a very important property distinguishing left and right operators of composition. Namely, the operator of right composition $\triangleright$ can be substituted by an operator $\textcircled{\triangleright}_K$, depending on a set $K$, simultaneously with changing the ordering of operations. Let us mention that no analogous operator exists for the operator of left composition $\triangleleft$.

**Theorem 1.** If $P_1$, $P_2$ and $P_3$ are such that $P_1 \triangleright P_2 \triangleright P_3$ is defined then

$$P_1 \triangleright P_2 \triangleright P_3 = P_1 \triangleright (P_2 \textcircled{\triangleright}_{K_1} P_3) = P_2 \textcircled{\triangleright}_{K_1} P_3 \triangleleft P_1,$$

where

$$P_2 \textcircled{\triangleright}_{K_1} P_3 = (P_3^{((K_1 \setminus K_2) \cap K_3)} P_2) \triangleright P_3.$$

*Proof.*
As $(P_3^{((K_1 \setminus K_2) \cap K_3)} P_2) \in \Pi^{(K_2 \cup (K_1 \cap K_3))}$

$$\begin{aligned}(P_3^{((K_1 \setminus K_2) \cap K_3)} P_2) \triangleright P_3 &= \frac{P_3^{((K_1 \setminus K_2) \cap K_3)} P_2 P_3}{P_3^{(K_3 \cap (K_2 \cup (K_1 \cap K_3)))}} \\ &= \frac{P_3^{((K_1 \setminus K_2) \cap K_3)} P_2 P_3}{P_3^{(K_3 \cap (K_1 \cup K_2))}}\end{aligned}$$

is a probability measure on $\mathbf{X}_{K_3 \cup K_2 \cup (K_1 \cap K_3)} = \mathbf{X}_{K_3 \cup K_2}$. In the following computations of its marginal measure on $\mathbf{X}_{K_1 \cap (K_2 \cup K_3)}$ we have to sum over $\mathbf{X}_{(K_2 \cup K_3) \setminus K_1}$. Therefore, $y^{(K)}$ for $y \in \mathbf{X}_{K_2 \cup K_3}$ will denote its projection into $\mathbf{X}_K$. Analogous meaning are those of the symbols $x^{(K)}$ and $z^{(K)}$.

For each $y \in \mathbf{X}_{K_2 \cup K_3}$

$$(P_2 \circledcirc_{K_1} P_3)^{(K_1 \cap (K_2 \cup K_3))}(y)$$

$$= \left( (P_3^{((K_1 \setminus K_2) \cap K_3)}(y^{((K_1 \setminus K_2) \cap K_3)}) P_2(y^{(K_2)})) \vartriangleright P_3(y^{(K_3)}) \right)^{(K_1 \cap (K_2 \cup K_3))}$$

$$= P_3^{((K_1 \setminus K_2) \cap K_3)}(y^{((K_1 \setminus K_2) \cap K_3)})$$
$$\cdot \sum_{x \in \mathbf{X}_{(K_2 \cup K_3) \setminus K_1}} \left[ P_3(x^{(K_3 \setminus K_1)} y^{(K_1 \cap K_3)}) \right.$$
$$\left. \cdot \frac{P_2(x^{(K_2 \setminus K_1)} y^{(K_1 \cap K_2)})}{P_3^{(K_3 \cap (K_1 \cup K_2))}(x^{((K_3 \cap K_2) \setminus K_1)} y^{(K_1 \cap K_3)})} \right]$$

$$= P_3^{((K_1 \setminus K_2) \cap K_3)}(y^{((K_1 \setminus K_2) \cap K_3)})$$
$$\cdot \sum_{x \in \mathbf{X}_{K_2 \setminus K_1}} \left[ P_2(x^{(K_2 \setminus K_1)} y^{(K_1 \cap K_2)}) \right.$$
$$\cdot \frac{1}{P_3^{(K_3 \cap (K_1 \cup K_2))}(x^{((K_3 \cap K_2) \setminus K_1)} y^{(K_1 \cap K_3)})}$$
$$\left. \cdot \left( \sum_{z \in \mathbf{X}_{K_3 \setminus (K_1 \cup K_2)}} P_3(x^{((K_2 \cap K_3) \setminus K_1)} z^{(K_3 \setminus (K_1 \cup K_2))} y^{(K_1 \cap K_3)}) \right) \right]$$

$$= P_3^{((K_1 \setminus K_2) \cap K_3)}(y^{((K_1 \setminus K_2) \cap K_3)})$$
$$\cdot \sum_{x \in \mathbf{X}_{K_2 \setminus K_1}} \left[ P_2(x^{(K_2 \setminus K_1)} y^{(K_1 \cap K_2)}) \right.$$
$$\cdot \frac{1}{P_3^{(K_3 \cap (K_1 \cup K_2))}(x^{((K_3 \cap K_2) \setminus K_1)} y^{(K_1 \cap K_3)})}$$
$$\left. \cdot P_3^{(K_3 \cap (K_1 \cup K_2))}(x^{((K_2 \cap K_3) \setminus K_1)} y^{(K_1 \cap K_3)}) \right]$$

$$= P_3^{((K_1 \setminus K_2) \cap K_3)}(y^{((K_1 \setminus K_2) \cap K_3)})$$
$$\cdot P_2^{(K_1 \cap K_2)}(y^{(K_1 \cap K_2)})$$

And thus

$$P_1 \vartriangleright (P_2 \circledcirc_{K_1} P_3) = \frac{P_1 \frac{P_3^{((K_1 \setminus K_2) \cap K_3)} P_2 P_3}{P_3^{(K_3 \cap (K_1 \cup K_2))}}}{(P_2 \circledcirc_{K_1} P_3)^{(K_1 \cap (K_2 \cup K_3))}}$$

$$= \frac{P_1 P_2 P_3 \frac{P_3^{((K_1 \setminus K_2) \cap K_3)}}{P_3^{(K_3 \cap (K_1 \cup K_2))}}}{P_3^{((K_1 \setminus K_2) \cap K_3)} P_2^{(K_1 \cap K_2)}}$$

$$= \frac{P_1 P_2 P_3}{P_3^{(K_3 \cap (K_1 \cup K_2))} P_2^{(K_1 \cap K_2)}}$$

$$= P_1 \vartriangleright P_2 \vartriangleright P_3,$$

which finishes the proof. $\square$

**Corollary 1.** If $K_2 \supseteq (K_1 \cap K_3)$ then

$$P_2 \circledcirc_{K_1} P_3 = P_2 \vartriangleright P_3$$

and thus also

$$P_1 \vartriangleright P_2 \vartriangleright P_3 = P_2 \vartriangleright P_3 \vartriangleleft P_1.$$

*Proof.*
The assertion is an immediate consequence of the fact

$$K_2 \supseteq (K_1 \cap K_3) \implies (K_1 \setminus K_2) \cap K_3 = \emptyset.$$
$\square$

**Lemma 5.** If $P_1$ and $P_2$ are consistent then

$$K_2 \supseteq (K_1 \cap K_3) \implies P_1 \vartriangleright P_2 \vartriangleright P_3 = P_1 \vartriangleright P_3 \vartriangleleft P_2.$$

*Proof.*

$$P_1 \vartriangleright P_3 \vartriangleleft P_2 = \frac{P_1 P_2 P_3}{P_3^{(K_1 \cap K_3)}(P_1 \vartriangleright P_3)^{(K_2 \cap (K_1 \cup K_3))}}.$$

Since we assume $K_2 \supseteq (K_1 \cap K_3)$

$$(P_1 \vartriangleright P_3)^{(K_2 \cap (K_1 \cup K_3))} = \frac{P_1^{(K_2 \cap K_1)} P_3^{(K_3 \cap K_2)}}{P_3^{(K_1 \cap K_3)}}$$

and therefore

$$P_1 \vartriangleright P_3 \vartriangleleft P_2 = \frac{P_1 P_2 P_3}{P_3^{(K_1 \cap K_3)} \frac{P_1^{(K_2 \cap K_1)} P_3^{(K_3 \cap K_2)}}{P_3^{(K_1 \cap K_3)}}}$$

$$= P_1 \vartriangleright P_2 \vartriangleright P_3$$

because, according to the assumptions, $P_1^{(K_1 \cap K_2)} = P_2^{(K_1 \cap K_2)}$ and $K_3 \cap (K_1 \cup K_2) = K_3 \cap K_2$. $\square$

**Lemma 6.** If $P_1$ and $P_3$ are consistent then

$$K_1 \supseteq (K_2 \cap K_3) \implies P_1 \vartriangleright P_2 \vartriangleright P_3 = P_1 \vartriangleright P_2 \vartriangleleft P_3.$$

*Proof.*
Since
$$P_1 \triangleright P_2 \triangleright P_3 = \frac{P_1 P_2 P_3}{P_2^{(K_1 \cap K_2)} P_3^{(K_3 \cap (K_1 \cup K_2))}},$$
and
$$P_1 \triangleright P_2 \triangleleft P_3 = \frac{P_1 P_2 P_3}{P_2^{(K_1 \cap K_2)} (P_1 \triangleright P_2)^{(K_3 \cap (K_1 \cup K_2))}}$$

these two expressions are equivalent because under the given assumptions
$$K_3 \cap (K_1 \cup K_2) = K_3 \cap K_1,$$
and $(P_1 \triangleright P_2)^{(K_1)} = P_1$ is consistent with $P_3$. □

The following Lemma is the only assertion presented in this paper expressing a condition allowing to change the ordering of operations of composition that does not require a special form of sets $K_i$. It is based only on a special form of conditional independence of measures $P_2$ and $P_3$ and coincidence of respective conditional measures.

**Lemma 7.** If
$$P_2^{(K_2 \cap K_3)} P_2^{(K_2 \cap K_1)} = P_2^{(K_2 \cap (K_1 \cup K_3))} P_2^{(K_1 \cap K_2 \cap K_3)},$$
$$P_3^{(K_3 \cap K_2)} P_3^{(K_3 \cap K_1)} = P_3^{(K_3 \cap (K_1 \cup K_2))} P_3^{(K_1 \cap K_2 \cap K_3)},$$
and
$$\frac{P_2^{(K_2 \cap K_3)}}{P_2^{(K_1 \cap K_2 \cap K_3)}} = \frac{P_3^{(K_3 \cap K_2)}}{P_3^{(K_1 \cap K_2 \cap K_3)}}$$
then
$$P_1 \triangleright P_2 \triangleright P_3 = P_1 \triangleright P_3 \triangleright P_2.$$

*Proof.*
$$P_1 \triangleright P_2 \triangleright P_3 = \frac{P_1 P_2 P_3}{P_2^{(K_2 \cap K_1)} P_3^{(K_3 \cap (K_1 \cup K_2))}}$$
and
$$P_1 \triangleright P_3 \triangleright P_2 = \frac{P_1 P_2 P_3}{P_3^{(K_3 \cap K_1)} P_2^{(K_2 \cap (K_1 \cup K_3))}}.$$

The denominators in both expressions equal each other as
$$P_2^{(K_2 \cap K_1)} P_3^{(K_3 \cap (K_1 \cup K_2))}$$
$$P_2^{(K_2 \cap K_1)} P_3^{(K_3 \cap K_1)} \frac{P_3^{(K_3 \cap K_2)}}{P_3^{(K_1 \cap K_2 \cap K_3)}}$$
$$= P_3^{(K_3 \cap K_1)} P_2^{(K_2 \cap K_1)} \frac{P_2^{(K_2 \cap K_3)}}{P_2^{(K_1 \cap K_2 \cap K_3)}}$$
$$P_3^{(K_3 \cap K_1)} P_2^{(K_2 \cap (K_1 \cup K_3))}.$$

Therefore if one of the expressions $P_1 \triangleright P_2 \triangleright P_3$, $P_1 \triangleright P_3 \triangleright P_2$ is defined then the other one must be defined, too, and then
$$P_1 \triangleright P_2 \triangleright P_3 = P_1 \triangleright P_3 \triangleright P_2.$$ □

**Corollary 2.** If $P_2$ and $P_3$ are consistent and for both $i = 2, 3$
$$P_i^{(K_i \cap (K_1 \cup K_{5-i}))} = P_i^{((K_i \cap K_1) \setminus K_{5-i})} P_i^{(K_2 \cap K_3)},$$
then
$$P_1 \triangleright P_2 \triangleright P_3 = P_1 \triangleright P_3 \triangleright P_2.$$

*Proof.*
We shall only verify that all the assumptions of the preceding Lemma are fulfilled.
$$\frac{P_2^{(K_2 \cap K_3)}}{P_2^{(K_1 \cap K_2 \cap K_3)}} = \frac{P_3^{(K_3 \cap K_2)}}{P_3^{(K_1 \cap K_2 \cap K_3)}}$$

follows immediately from the consistency of $P_2, P_3$.

Moreover, under the given assumptions
$$P_i^{(K_i \cap K_1)} = (P_i^{(K_i \cap (K_1 \cup K_{5-i}))})^{(K_i \cap K_1)}$$
$$= (P_i^{((K_i \cap K_1) \setminus K_{5-i})} P_i^{(K_2 \cap K_3)})^{(K_i \cap K_1)}$$
$$= P_i^{((K_i \cap K_1) \setminus K_{5-i})} P_i^{(K_1 \cap K_2 \cap K_3)}$$

and therefore
$$P_i^{(K_i \cap K_1)} P_i^{(K_2 \cap K_3)}$$
$$= P_i^{((K_i \cap K_1) \setminus K_{5-i})} P_i^{(K_1 \cap K_2 \cap K_3)} P_i^{(K_2 \cap K_3)}$$
$$= P_i^{(K_i \cap (K_1 \cup K_{5-i}))} P_i^{(K_1 \cap K_2 \cap K_3)}$$

for both $i = 2, 3$. □

## 4  GRAPH MODELS

This section should give a hint to answering the question why we were so much interested in finding conditions under which changing an ordering of the operations of composition does not influence the resulting measure. We shall assume that $P_i \in \Pi^{(K_i)}$ for $i = 1, \ldots, n$, and, moreover, that
$$\bigcup_{i=1}^{n} K_i = V.$$

The reader familiar with Bayesian networks and decomposable models can immediately see that both these probabilistic models can be expressed in our notation as
$$P_1 \triangleright P_2 \triangleright \ldots \triangleright P_n.$$

An application of such a probabilistic model $Q = P_1 \triangleright P_2 \triangleright \ldots \triangleright P_n$ to inference, i.e. for computation of a (say) conditional probability $Q(X_r = a | X_s = b)$ is simple if, for example, $s \in K_1$, $r \in K_m$ and no $K_{m+1}, K_{m+2}, \ldots, K_n$ contains any of the indices $r$ and $s$. On the contrary, however, the computational procedure can become extremely complex if $s \in K_n$ and $r \in K_m$ for some $m \neq n$. Therefore, sequencies in which the ordering of the oligo-dimensional measures can be changed without changing the resulting multi-dimensional measure are of great importance. Of great importance are also the perfect sequences introduced below.

We shall say that an ordered sequence of probability measures $P_1, \ldots, P_n$ is *perfect* if

$$P_1 \triangleright P_2 = P_1 \triangleleft P_2,$$
$$P_1 \triangleright P_2 \triangleright P_3 = P_1 \triangleleft P_2 \triangleleft P_3,$$
$$\vdots$$
$$P_1 \triangleright \ldots \triangleright P_n = P_1 \triangleleft \ldots \triangleleft P_n.$$

For these sequences the following two simple assertions hold. The first one is an immediate consequence of an inductive application of Lemma 3.

**Lemma 8.** Let $P_1 \triangleright P_2 \triangleright \ldots \triangleright P_n$ be defined. The sequence $P_1, P_2, \ldots, P_n$ is perfect if and only if the pair of measures $(P_1 \triangleright \ldots \triangleright P_{m-1})$ and $P_m$ are consistent for all $m = 2, 3, \ldots, n$. □

**Theorem 2.** Let $P_1, P_2, \ldots, P_n$ be a perfect sequence of oligo-dimensional probability measures. Then

1. $P_1 \triangleright P_2 \triangleright \ldots \triangleright P_n \in \bigcap_{i=1}^{n} \Pi^{(V)}(P_i)$, and

2. $P_1 \triangleright P_2 \triangleright, \ldots, \triangleright P_n \in \Pi^{(V)}(P_1 \triangleright \ldots \triangleright P_m)$ for all $m = 1, \ldots, n-1$.

*Proof.*
Consider any $m \in \{1, \ldots, n-1\}$ and denote $Q = P_1 \triangleright \ldots \triangleright P_m$. All distributions

$$Q \triangleright P_{m+1},$$
$$Q \triangleright P_{m+1} \triangleright P_{m+2},$$
$$\vdots$$
$$Q \triangleright P_{m+1} \triangleright \ldots \triangleright P_n$$

have $Q$ for their marginal measure and thus $P_1 \triangleright \ldots \triangleright P_n = Q \triangleright P_{m+1} \triangleright \ldots, \triangleright P_n \in \Pi^{(V)}(Q)$ which proves the assertion 2.

Since, according to the perfectness of $P_1, \ldots, P_m$,

$$Q = P_1 \triangleleft \ldots \triangleleft P_m,$$

it is clear that

$$Q \in \Pi^{(K)}(P_m),$$

where $K = K_1 \cup \ldots \cup K_m$. From $P_1 \triangleright \ldots \triangleright P_n \in \Pi^{(V)}(Q)$ and $Q \in \Pi^{(K)}(P_m)$, we get immediately $P_1 \triangleright \ldots \triangleright P_n \in \Pi^{(V)}(P_m)$. Since $m \leq n-1$ was arbitrary and $P_1 \triangleright \ldots \triangleright P_n \in \Pi^{(V)}(P_n)$ follows from the perfectness condition, we have finished the first part of the proof, too. □

Thus, like for Bayesian networks and decomposable models, marginalization for perfect sequences can be done by a simple neglect of a "tail" of the sequence. Moreover, the direct consequence of the preceding assertion is the fact that each Bayesian network

$$P_1 \triangleright \ldots \triangleright P_n$$

can be transformed into a perfect sequence.

**Corollary 3.** If $P_1 \triangleright \ldots \triangleright P_n$ is defined then the sequence $Q_1, \ldots, Q_n$ computed by the following process

$$Q_1 = P_1,$$
$$Q_2 = Q_1^{(K_2 \cap K_1)} \triangleright P_2,$$
$$Q_3 = Q_2^{(K_3 \cap (K_1 \cup K_2))} \triangleright P_3,$$
$$\vdots$$
$$Q_n = Q_{n-1}^{(K_n \cap (K_1 \cup \ldots K_{n-1}))} \triangleright P_n$$

is perfect and

$$P_1 \triangleright \ldots \triangleright P_n = Q_1 \triangleright \ldots \triangleright Q_n. \qquad \Box$$

From a classical result of Kellerer (Kellerer 1964) it follows almost directly that for a sequence of pairwise consistent measures $P_1, \ldots, P_n$ such that the sequence $K_1, K_2, \ldots, K_n$ meets the *running intersection property*:

$$\forall i = 3, \ldots, n \ \exists j \ (1 \leq j < i) \ \left( (K_i \cap (\bigcup_{k=1}^{i-1})) \subseteq K_j \right),$$

the sequence $P_1, \ldots, P_n$, defining a decomposable model

$$P_1 \triangleright \ldots \triangleright P_n,$$

is perfect, too.

In the abstract, we expressed our conviction that most of theoretical results and procedures connected with Bayesian networks and decomposable models can be translated into the terminology of perfect sequences. To support this statement, let us conclude the paper by describing the procedure introduced by Lauritzen and Spiegelhalter (Lauritzen, Spiegelhalter 1988) and known under the name of *local computations*. This procedure is, in fact, nothing else than a transformation of a Bayesian network into a decomposable model.

**Procedure**

Let $P_1, P_2, \ldots, P_n$ be a perfect sequence of oligo-dimensional measures and $(L_1, L_2, \ldots, L_m)$ be an arbitrary *decomposable covering* of V (i.e. $L_1, \ldots, L_m$ meets the running intersection property and $(L_1 \cup \ldots \cup L_m) = V$) such that

$$\forall i = 1, \ldots, n \ \exists j \in \{1, \ldots, m\} \quad (K_i \subseteq L_j).$$

The index $j$, existence of which is guaranteed by the above condition, will be denoted by $b(i)$ in the sequel (if there are more than one such $j$, $b(i)$ is an arbitrary of them). Compute $m \times n$ oligo-dimensional measures $R_{(1,1)}, \ldots, R_{(1,m)}, R_{(2,1)}, \ldots, R_{(n,m)}$. For $i = 1$ compute

$$R_{(1,j)} = P_1^{(L_j \cap K_1)} \quad \text{for all } j = 1, \ldots, m.$$

For $i > 1$ compute

$$R_{(i,b(i))} = R_{(i-1,b(i))} \triangleright P_i.$$

To compute $R_{(i,j)}$ for $i > 1$ and $j \neq b(i)$ find an ordering

$$j_1 = b(i), j_2, \ldots, j_m$$

such that $L_{j_1}, \ldots, L_{j_m}$ meets the running intersection property (such an ordering always exists). Then for each $k > 1$ there exists $\ell < k$ for which

$$(L_{j_k} \cap (L_{j_1} \cup \ldots \cup L_{j_{k-1}})) \subseteq L_{j_\ell}$$

and then $R_{(i,j_k)}$ can be computed

$$R_{(i,j_k)} = R_{(i-1,j_k)} \triangleleft R_{(i,j_\ell)}^{(L_{j_k} \cap L_{j_\ell})}.$$

◇

**Theorem 3.** *Let a perfect sequence $P_1, P_2, \ldots, P_n$, a decomposable covering $L_1, \ldots, L_m$ of V, and measures $R_{(i,j)}$ ($i = 1, \ldots, n, j = 1, \ldots, m$) be as in Procedure.*

*Then the measures $R_{(i,j)}$ are consistent with the measure $P_1 \triangleright \ldots \triangleright P_i$ for all $i = 1, \ldots, n, j = 1, \ldots, m$.*

*Proof.*
When proving the assertion we shall proceed in the three steps: (a) for $i = 1$, (b) for $i > 1$ and $j = b(i)$, and (c) for remaining indices $i, j$.

**(a)** The fact that all $R_{(1,j)}$ are consistent with $P_1$ follows directly from

$$R_{(1,j)} = P_1^{(L_j \cap K_1)}.$$

**(b)** Now, assume that the assertion holds for a fixed $i$, $(1 \leq i < n)$. Thus, $R_{(i,b(i+1))}$ must be consistent with $P_1 \triangleright \ldots \triangleright P_i$, which means, in this case,

$$(P_1 \triangleright \ldots \triangleright P_i)^{(L_{b(i+1)})} = R_{(i,b(i+1))}.$$

Since $L_{b(i+1)} \supseteq K_{i+1} \supseteq K_{i+1} \cap (K_1 \cup \ldots \cup K_i)$ we get from Lemma 2

$$(P_1 \triangleright \ldots \triangleright P_{i+1})^{(L_{b(i+1)})} = R_{(i,b(i+1))} \triangleright P_{i+1}.$$

**(c)** To conclude the proof consider two indices $i$, $(1 \leq i < n)$ and $k$, $(1 < k \leq m)$ and assume that the assertion holds for all $R_{(i,j)}$, $j = 1, \ldots, m$ and also for $R_{(i+1,j_1)}, R_{(i+1,j_2)}, \ldots, R_{(i+1,j_{k-1})}$, where $j_1 = b(i), \ldots, j_m$ is the ordering from Procedure. We shall prove that it holds also for $R_{(i+1,j_k)}$.

In accordance with Procedure, denote $\ell$ that index for which

$$(L_{j_k} \cap (L_{j_1} \cup \ldots \cup L_{j_{k-1}})) \subseteq L_{j_\ell}.$$

Let us compute

$$(P_1 \triangleright \ldots \triangleright P_{i+1})^{(L_{j_k})}$$
$$= (P_1 \triangleright \ldots \triangleright P_i)^{(L_{j_k})} \triangleleft (P_1 \triangleright \ldots \triangleright P_{i+1})^{(L_{j_k} \cap K)}$$
$$= R_{(i,j_k)} \triangleleft (P_1 \triangleright \ldots \triangleright P_{i+1})^{(L_{j_k} \cap K)}$$

for any $K$ such that $K \cap L_{j_k} \supseteq K_{i+1} \cap L_{j_k}$.

As

$$L_{j_k} \cap K_{i+1} \subseteq L_{j_k} \cap L_{j_1} \subseteq L_{j_k} \cap (L_{j_1} \cup \ldots \cup L_{j_{k-1}})$$
$$\subseteq L_{j_k} \cap L_{j_\ell}$$

we can take $K = L_{j_k} \cap L_{j_\ell}$ in the last equality. Since we assume that $R_{(i+1,j_\ell)}$ is consistent with $P_1 \triangleright \ldots \triangleright P_{i+1}$ we can thus substitute $(P_1 \triangleright \ldots \triangleright P_{i+1})^{(L_{j_k} \cap K)}$ by $R_{(i+1,j_\ell)}^{(L_{j_k} \cap L_{j_\ell})}$, which finishes the proof. □

**Theorem 4.** *Let a perfect sequence $P_1, P_2, \ldots, P_n$, a decomposable covering $L_1, \ldots, L_m$ of V, and measures*

$R_{(i,j)}$ $(i = 1, \ldots, n,\ j = 1, \ldots, m)$ be as in Procedure. Then $R_{(n,1)}, \ldots, R_{(n,m)}$ is a perfect sequence, too, and

$$R_{(n,1)} \triangleright \ldots \triangleright R_{(n,m)} = P_1 \triangleright \ldots \triangleright P_n.$$

*Proof.*
This assertion follows directly from the preceding Theorem and the fact (cf. e.g. Theorem 1.5.7 in (Hájek et al. 1992)) that if two measures can be expressed as products of functions defined on $\mathbf{X}_{L_1}, \ldots, \mathbf{X}_{L_m}$ and their marginals on these subspaces coincide then these measures are equivalent. □

## 5 CONCLUSIONS

We have presented a new apparatus based on the operators of left and right compositions. Though the operators are defined by almost identical formulae they substantially differ when used iteratively to compose sequences of probabilistic measures. The difference consists mainly in computational complexity of the respective processes. In contrast to the operator of right composition ▷, iterative application of the operator of left composition ◁ rather often leads to computationally intractable procedures because of increasing dimensionality of the measures from which the normalization factors are computed. Another noteworthy difference is an existence of the operator $\triangleright_K$ guaranteed by Theorem 1. Let us mention that the existence of this operator is, in a sense, closely connected with ideas of Bayesian network marginalization (by nodes deletion and edges reversal) proposed by R. Shachter (Shachter 1986a, 1986b, 1988).

The probabilistic models described by perfect sequences of (oligo-dimensional) probability measures embody Bayesian networks (and, naturally, decomposable models, too). We conjecture, that these models constitute the very class of models for which effective computational procedures can be found. Therefore, in the future, we shall concentrate to algoritmization of these processes for which the theoretical background is given by assertions in section 3.

### Acknowledgments

This work was supported by the grant of Ministry of Education of ČR no. VS96008.